# Normalized Cut Loss for Weakly-supervised CNN Segmentation

Meng Tang[2,1][*]    Abdelaziz Djelouah[1]    Federico Perazzi[1]
Yuri Boykov[2]    Christopher Schroers[1]

[1]Disney Research, Zürich, Switzerland
[2]Computer Science, University of Waterloo, Canada

## Abstract

*Most recent semantic segmentation methods train deep convolutional neural networks with fully annotated masks requiring pixel-accuracy for good quality training. Common weakly-supervised approaches generate full masks from partial input (e.g. scribbles or seeds) using standard interactive segmentation methods as preprocessing. But, errors in such masks result in poorer training since standard loss functions (e.g. cross-entropy) do not distinguish seeds from potentially mislabeled other pixels. Inspired by the general ideas in semi-supervised learning, we address these problems via a new principled loss function evaluating network output with criteria standard in "shallow" segmentation, e.g. normalized cut. Unlike prior work, the cross entropy part of our loss evaluates only seeds where labels are known while normalized cut softly evaluates consistency of all pixels. We focus on normalized cut loss where dense Gaussian kernel is efficiently implemented in linear time by fast Bilateral filtering. Our normalized cut loss approach to segmentation brings the quality of weakly-supervised training significantly closer to fully supervised methods.*

## 1. Introduction

Since the seminal work [29], deep convolutional neural networks (CNN) dominate almost all aspects of computer vision, e.g. recognition [45, 25], detection [20, 41], and segmentation [33, 13]. It is capable of learning intermediate representations at different levels given abundant training data. For semantic segmentation, all leading methods in PASCAL VOC 2012 train some *fully convolutional networks* (FCN) based on given ground-truth segmentations. Typically, pixel-wise cross entropy loss is minimized.

Supervised training of FCNs requires a huge number of fully annotated ground-truth masks that is costly to obtain. Training with weak annotations, e.g. scribbles [32, 51], bounding boxes [26, 16, 51, 36], clicks [4], and image-level tags [51, 36], has caught a lot of interest recently.

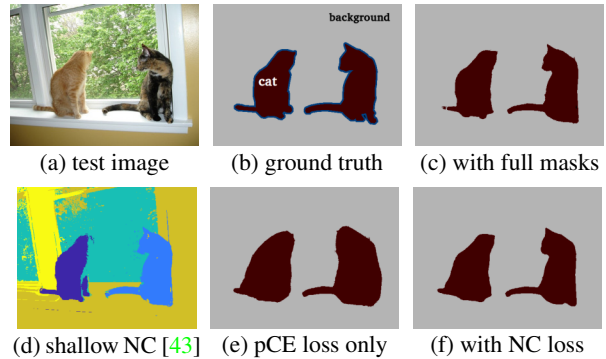

(a) test image    (b) ground truth    (c) with full masks

(d) shallow NC [43]    (e) pCE loss only    (f) with NC loss

Figure 1: Unsupervised *normalized cut* (NC) segmentation exploits low-level similarities (d). Weak supervision using partial cross entropy (pCE) on scribbles with our NC loss trains CNN (f) nearly as good as full mask supervision (c).

Common approaches often use standard *shallow*[1] segmentation techniques (e.g. graph cuts [9, 42, 31]) with seeds or boxes to generate <u>fake</u> full masks (proposals) to be used for training, which is often iterated with proposal generation [32, 26, 36]. However, inaccuracies of such masks (segmentation proposals) mislead training since typical cross entropy (CE) loss is minimized over mislabeled points and the network over-fits mistakes, see Sec.3.2. According to semi-supervised learning literature [55, 12], early mistakes reinforce themselves in self-learning. As we show, better training can be achieved by minimizing *partial* cross entropy (pCE) loss on <u>true</u> scribbles only, see Fig.1(e).

Our approach to weakly-supervised CNN training is motivated both by common ideas in semi-supervised learning and by standard criteria in shallow image segmentation. In contrast to existing proposal generating technqiues, we advocate a principled yet simple and general approach: *regularized* semi-supervised loss directly integrating shallow segmentation criteria into CNN training, see Fig.1(d,f). This paper focuses on a popular balanced segmentation criteria - *normalized cut* [43]. Our main contributions are:

---

[1]Here "shallow" refers to techniques unrelated to neural networks.

- We propose and evaluate a novel loss for weakly supervised semantic segmentation. It combines partial cross entropy on labeled pixels and *normalized cut* for unlabeled pixels. Losses based on CRF and other shallow regularizers are studied in our follow-up work [47].

- Even without normalized cut loss, our partial cross entropy loss for scribbles (*loss sampling*) works surprisingly well compared to cross entropy over "generated" full masks that make the network over-fit to mistakes.

- We show efficient implementation for normalized cut loss layer with dense Gaussian kernel in linear time.

- Experiments show that normalized cut loss achieves the state-of-the-art for training semantic segmentation with scribbles. We evaluate other losses in [47].

## 2. Background and Motivation

Our regularized loss is inspired by the general ideas for semi-supervised (deep) learning [12, 50, 7] and by the standard regularization objectives in 'shallow' image segmentation or clustering, as discussed below.

**Regularized semi-supervised losses:** Various forms of regularization are widely used in machine learning and neural networks in particular, see Sec.3.3 for an overview. This paper is focused specifically on regularized losses for semi-supervised learning with partially labeled training data. In this case regularization is directly applied to the network output [50, 7] rather than to the network parameters. Typical regularized semi-supervised loss function over the network output combines two terms

- *Fidelity* of network output to the labeled data
- *Regularization* of the entire network output

The purpose of the regularization term is to propagate the empirical losses (fidelity) over partially labeled input (mask) to the entire training data including unlabeled points.

In particular, Weston [50] proposed a general idea that a loss for network's output can incorporate regularizers from standard "shallow" semi-supervised methods. Assume that variables $S_p \in [0,1]^K$ describe the network's output for $p \in \Omega$. Then, one loss function example in [50] based on a common *Laplacian eigenmaps* regularizer [6] can be written as

$$\sum_{p \in \Omega_{\mathcal{L}}} \ell(S_p, y_p) \;+\; \lambda \sum_{p,q \in \Omega} W_{pq} \, \|S_p - S_q\|^2 \quad (1)$$

where $l$ is any standard loss for labeled points $p \in \Omega_{\mathcal{L}}$ with known ground truth labels $y_p$. The regularization term above softly enforces output consistency among all points based on predefined pairwise affinities $W = [W_{pq}]$.

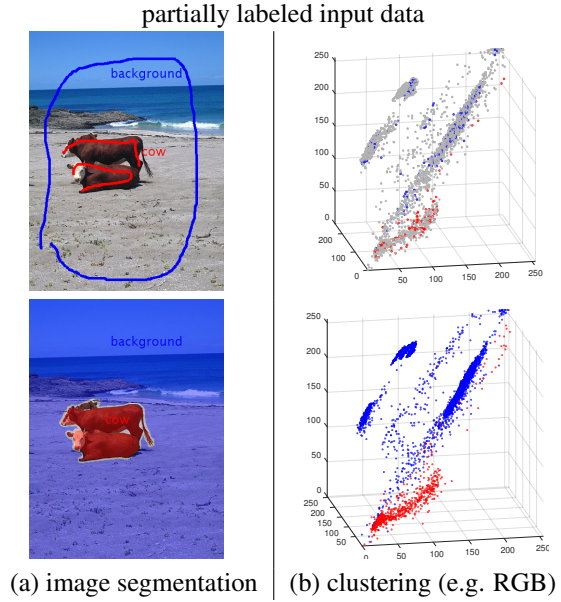

(a) image segmentation | (b) clustering (e.g. RGB)

Figure 2: Interactive segmentation (a) and (transductive) semi-supervised "shallow" learning (b) are well-known related problems, e.g. both of them could be formulated via similar MRF/CRF regularizers like (2) in [9, 55] or *normalized cut* [43]. The use of shallow learning regularizers in semi-supervised losses on DNN outputs [50] motivates direct integration of standard image segmentation regularizers into losses for weakly-supervised CNN segmentation.

**Shallow semi-supervised segmentation & clustering:** Motivated by the general ideas above, we propose to incorporate regularizers from "shallow" segmentation/clustering into CNN segmentation loss. Due to importance of low-level regularizers for segmentation and computer vision in general, there are many choices available in the literature [19, 11, 9, 43, 23, 34, 39, 30, 15, 46]. To further motivate our approach, we discuss one specific basic example of such regularizer.

Probably the simplest energy for shallow interactive image segmentation in [9] that also works for transductive semi-supervised learning (clustering) [55] combines hard constraints over seeds $p \in \Omega_{\mathcal{L}}$ with predefined labels $y_p$ and the Potts regularizer

$$\sum_{p \in \Omega_{\mathcal{L}}} -\log S_p^{y_p} \;+\; \lambda \sum_{p,q \in \Omega} W_{pq} \, [S_p \neq S_q] \quad (2)$$

where $S_p^k \in \{0,1\}$ are now interpreted as class assignment indicators[2] and $[\cdot]$ are Iverson brackets. The first term enforces $S_p^{y_p} = 1$ for $p \in \Omega_{\mathcal{L}}$. The regularization term penalizes disagreements between the pairs of points. Figure 2 illustrates shallow interactive image segmentation (a) and semi-supervised clustering of general data (b).

---

[2]This is only a minor conflict with earlier notation $S_p^k \in [0,1]$ for real-valued network output; in the context of CNN segmentation such relaxed output variables $S_p^k$ also correspond to (soft-max) class assignments.

Properties of the regularization term in (2) heavily depend on weights $[W_{pq}]$. For sparse non-negative matrix $W$ with non-zero weights concentrated on nearest neighbors as in classical Ising or Potts models, there is a geometric interpretation corresponding to the weighted *length* of the segmentation boundary [10]. For denser $W$ the regularizer in (2) reduces to a quadratic function over weighted cardinalities of segments[3]. For dense Gaussian affinity, the regularizer above corresponds to a *dense CRF* [28]. Instead of pairwise regularizer in (2), later Sections 3.4 and 4 focus on *normalized cut* objective [43], which is a popular balanced segmentation and unsupervised clustering criterion.

**Connecting shallow and deep segmentation:** Following the general idea of integrating (shallow) regularizers into semi-supervised loss for (deep) learning [12, 50, 7], we advocate this principled approach in the context of weakly-supervised CNN segmentation. That is, we propose semi-supervised training loss on CNN output to combine empirical risk (e.g. cross entropy) over labeled pixels and a regularizer on all pixels common in "shallow" segmentation.

In order to integrate common segmentation energies like (2) into CNN loss functions, they should have a "relaxed" formulation extendable to real-valued segmentation variables $S_p \in [0,1]^K$ (on probability simplex) typically produced by soft-max output of the network. For example, the Potts or dense CRF term in (2) can be written as a quadratic form based on linear algebraic notation giving an equivalent formulation of shallow segmentation energy (2)

$$\sum_{p \in \Omega_{\mathcal{L}}} -\log S_p^{y_p} \;+\; \lambda \sum_k S^{k'} W (\mathbf{1} - S^k) \qquad (3)$$

where $S^k \in [0,1]^{|\Omega|}$ is a (soft) support vectors for class $k$ combining $k$-th components $S_p^k$ of vectors $S_p \in [0,1]^K$ for all points $p \in \Omega$. Interestingly, the first term in (2) representing hard constraints over seeds for binary indicators $S_p \in \{0,1\}^K$ becomes a *partial cross entropy* term in case of relaxed variables $S_p \in [0,1]^K$.

Equation (3) is a basic example of segmentation energy that could be used as *regularized semi-supervised loss* directly over real-valued CNN output. This fits previously discussed ideas for regularized semi-supervised losses for learning in general. For example, loss (3) is very closely related to (1) where the second term is also a quadratic relaxation of the Potts model in (2) different from (3).

In general, regularized semi-supervised loss functions for CNN segmentation can use differentiable relaxed versions of many standard shallow segmentation regularizers such as Potts [9], dense CRF [28], or their combinations with balanced clustering criteria [46]. We will study these losses in the future. This paper focuses on normalized cut objective [43]. As discussed in Section 3.4, it differs from the Potts and denseCRF models by normalization that encourages segment balancing and addresses shrinking bias. This objective also allows a continuous relaxation [46].

---

[3]Extreme $W_{pq} = const$ gives a negative sum of squared cardinalities.

While our future work plans include an empirical comparison for all these models in the context of regularized weakly-supervised losses for CNN segmentation, this paper focuses specifically on the *normalized cut loss* thoroughly discussed in Sec. 3.4 and 4.2.

## 3. Related Work

### 3.1. Semi-supervised Learning

Semi-supervised learning is about learning from both labeled and unlabeled data [55, 12]. Graph based algorithms [8, 54, 53, 6] are the most relevant to our work. It assumes that two nodes with larger graph affinity are more likely to have the same label. Graph Cut [8] solves a combinatorial problem in polynomial time. Harmonic Functions [54] relaxes discrete labeling to real values and admits a closed-form solution. Manifold regularization [6] prevents over-fitting to training examples by including extra regularization on the whole feature space. As such, it gives better generalization and a natural extension to test data.

Semi-supervised learning ideas with "shallow" models are adjusted into deep learning in the general framework by Weston *et al.* [50]. Extra losses based on semi-supervised embedding are minimized by standard back propagation. Indeed, our regularized loss is similar to the scheme in [50] regularizing network output. However, we are the first to utilize ideas in such principled framework for weakly-supervised CNN segmentation.

### 3.2. Weakly-supervised Semantic Segmentation

Semantic segmentation has been addressed with scribbles [32, 51, 48], bounding boxes [26, 16, 51, 36], clicks [4] or even image-level tags [51, 36, 27].

Xu *et al.* [51] formulated all types of weak supervision as linear constraints in max-margin clustering. This framework is rather flexible and principled. However, we take advantage of deep CNN rather than SVM used in [51].

Recent work [16, 32, 26, 36, 27, 38, 48] train CNNs from segmentation "proposals", which can be crude box labelings or the outputs of shallow interactive segmentation. Typically, such methods make (expensive) inference steps generating (fake) ground truth masks/proposals and then minimize cross-entropy w.r.t. such masks, potentially over-fitting to their errors. For example, ScribbleSup [16] iteratively generate proposals/masks via graph cuts, while [27] use additional CRF inference layers to produce them. Instead of MRF/CRF regularization, [38] uses cardinality constraints to generate explicit full proposals/masks.

With our joint loss, it suffices to train in one pass and we don't need iterative training and heuristic segmentation proposals. It is true that many off-line interactive segmentation algorithms exist [42, 31, 40]. However, segmentation proposals have major limitations discussed bellow.

**Why not train with segmentation proposal?** Mistakes reinforce themselves in self-learning scheme [12, 55]. Indeed, as also briefly mentioned in Sec. 3.1, self-learning is one of the earliest ideas for semi-supervised learning [12], which doesn't have any convergence guarantee.

In practice, "shallow" segmentation proposals are likely to be erroneous, see examples in Fig. 7. Most interactive segmentation methods don't consider semantic cue. As such the proposals are misleading for training. Instead of generating unreliable proposals and train models to fit errors, our method is more direct, incorporating standard segmentation regularizer as a loss. Also heuristic pre-processing is not favored in semi-supervised learning. Most modern semi-supervised learning approaches minimize a regularized loss with e.g. SVM or neural networks [6, 50].

### 3.3. Regularization for Neural Networks

Regularizations have also been widely used in neural networks to avoid over-fitting or encourage sparsity, e.g. Norm regularization [22], Dropout [44] and ReLU [21].

Our normalized cut loss differs from these regularization. Ours is a semi-supervised loss for regularizing network output for unlabeled data. Such regularization is well coupled with partial fidelity loss, allowing implicit label propagation during training.

Regularization techniques for CNN segmentation include post-processing (e.g. dense CRF [28, 13]), and appended trainable layers (e.g. CRF-RNN [52], Bilateral Solver [3]). Our regularized loss for weakly-supervised segmentation is very different. It's beyond the scope of this work to compare all schemes for regularization in fully- or weakly-supervised CNN segmentation.

### 3.4. Normalized Cut and Image Segmentation

Normalized Cut is a popular graph clustering algorithm originally proposed for image segmentation [43]. It is the sum of ratios between the cuts and the volumes.

$$\sum_k \frac{cut(\Omega_k, \Omega/\Omega_k)}{assoc(\Omega_k, \Omega)} \equiv \sum_k \frac{S^{k'}W(\mathbf{1}-S^k)}{d'S^k}, \quad (4)$$

where $\Omega_k$ is the set of pixels labeled $k$ and $S^k$ is binary indicator vector. The *cut* or *assoc* for two sets $A$ and $B$ is defined as $\sum_{p \in A, q \in B} W_{pq}$, see [43].

Normalized Cut is a variant of a family of spectral clustering and embedding algorithms [35, 5, 49] that typically depend on the eigenvectors of unnormalized or normalized Laplacian matrix. As a segmentation regularizer, normalized cut differs from Potts [9] and dense CRF [28] by having extra normalization. As such, it encourages balanced clustering and voids shrinking bias. So in this work, we focus on normalized cut for these appealing properties [43] and its popularity in segmentation.

## 4. Our Method

We propose a joint loss of (partial) cross entropy and normalized cut for weakly-supervised CNN segmentation. The partial cross entropy is briefly discussed in Sec. 4.1. Our main contribution, using normalized cut loss as a regularizer, is presented in Sec. 4.2. A fast implementation of our normalized cut loss layer with a dense Gaussian kernel is introduced in Sec. 4.3.

### 4.1. Partial Cross Entropy as Loss Sampling

The simple idea behind the partial cross entropy loss is to only consider the cross entropy loss for labeled pixels $p \in \Omega_\mathcal{L}$ which effectively ignores other regions. We are not the first to ignore regions in weakly-supervised segmentation in general, as there are examples for boxes [26] and for clicks [4]. However, this partial loss can be seen as a sampling of the loss with full masks by rewriting it in the following way:

$$\sum_{p \in \Omega_\mathcal{L}} -\log S_p^{y_p} = \sum_{p \in \Omega} -u_p \cdot \log S_p^{y_p}. \quad (5)$$

Here, $u_p = 1$ for $p \in \Omega_\mathcal{L}$ and 0 otherwise. We interpret $u_p$ as sampling on $\Omega$ for randomly drawn scribbles.

In practice, we found that training only with this simple loss works surprisingly well, achieving more than 85% of the accuracy compared to using the full labeling. In fact, using this loss is even better than training from GrabCut proposals as we show in our experiments in Sec. 5.2.1 but this trick has been overlooked in previous work [32]. This supports our argument in Sec. 3.2 that segmentation proposals may be very misleading.

### 4.2. Normalized Cut Loss

Given any affinity matrix $W = [W_{ij}]$ and degree vector $d = W\mathbf{1}$, we define our joint loss for one image as

$$\underbrace{\sum_{p \in \Omega_\mathcal{L}} -\log S_p^{y_p}}_{\text{(partial) Cross Entropy}} + \lambda \underbrace{\sum_k \frac{S^{k'}W(\mathbf{1}-S^k)}{d'S^k}}_{\text{(continuous) Normalized Cut}}. \quad (6)$$

The first term penalizes the partial cross entropy while the second term represents a (relaxed) normalized cut for relaxed segmentation $S^k \in [0,1]^{|\Omega|}$. We use standard Gaussian kernel $W_{ij}$ over RGBXY space. Below we further justify why normalized cut offers a proper regularization for semantic segmentation.

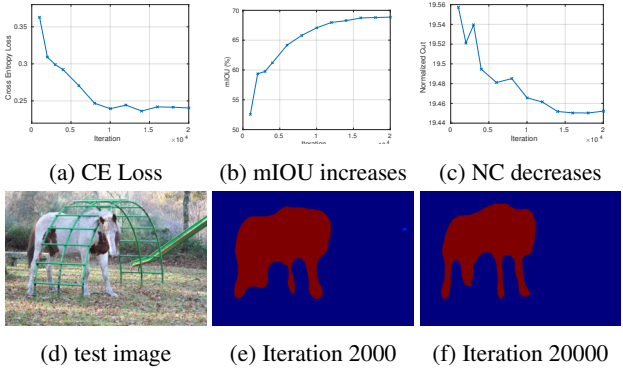

(a) CE Loss  (b) mIOU increases  (c) NC decreases

(d) test image  (e) Iteration 2000  (f) Iteration 20000

Figure 3: Statistics of val. set when training with masks. Cross entropy decreases as expected. Evaluated normalized cut decreases as mIOU increases. For the horse example, from iteration 2000 to 20000, mIOU improves from 86.2% to 92.4% while normalized cut decreases from 0.60 to 0.54. Good semantic segmentation gives low normalized cut.

With a simple Gaussian kernel on RGBXY, normalized cut produces clustering/segmentation that may not be semantically meaningful. However, in practice we observed that better semantic segmentations typically correspond to lower normalized cut energies, see Fig. 3. This suggests that normalized cut is a reasonable loss encouraging balanced non-linear partitioning of unlabeled pixels. Our normalized cut loss is motivated by popularity of normalized cut as an unsupervised segmentation criteria with many attractive properties [43, 49].

An important detail is that in our implementation of the normalized cut loss, $S_p$ for scribbles $p \in \Omega_{\mathcal{L}}$ is clamped to be their ground truth labeling. As such, the scribbles serve as seeds from which the ground truth labels are implicitly propagated to unknown pixels during training.

### 4.3. Gradient Computation

This section shows how to compute the gradient of the normalized cut loss regularizer with a dense Gaussian kernel in linear time. First, we rewrite the normalized cut using the equivalent normalized association:

$$E_{NC}(S) = \sum_k \frac{S^{k'}W(\mathbf{1}-S^k)}{d'S^k} \stackrel{c}{=} \sum_k -\frac{S^{k'}WS^k}{d'S^k}. \quad (7)$$

Its gradient w.r.t. $S^k$ is,

$$\frac{\partial E_{NC}(S)}{\partial S^k} = \frac{S^{k'}WS^k d}{(d'S^k)^2} - \frac{2WS^k}{d'S^k}. \quad (8)$$

Since we are assuming a dense Gaussian kernel, a naive implementation of forward and backward passes for the normalized cut loss is prohibitively slow ($O(|\Omega|^2)$). The bottleneck is to compute $d = W\mathbf{1}$ and $AS^k$. With a fixed width

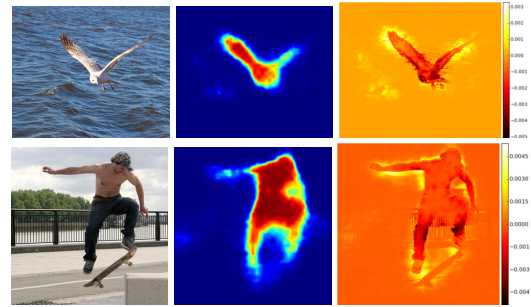

Figure 4: Network output (middle) and corresponding gradient (right) of normalized cut loss w.r.t. soft-max input. Training with such gradients gives better color clustering.

Gaussian kernel for 5-d RGBXY, this comes down to solving a bilateral filtering with many available fast computation techniques [37, 2, 1]. We use the permutohedral lattice [1] with linear time complexity in $|\Omega|$ and dimensions. Thus, each forward evaluation and back-propagation through the normalized cut loss is efficient.

Unlike a pixel-wise loss, our normalized cut loss is high-order and its gradients (8) are not intuitive. We show a visualization of the gradients in Sec. 5, see Fig. 4. The gradients indeed encourage better color clustering.

Interestingly, the same gradients appear as the slopes of linear upper bound [46] for normalized cut, which is proved to be concave with a PSD affinity matrix. This helps us to see that gradient descent for neural networks is likely to decrease the concave normalized cut loss.

## 5. Experiments

Our joint loss (6) combines partial cross entropy and normalized cut regularization. To see how capable are neural networks to minimize normalized cut, we train networks for *normalized cut loss only* in Sec. 5.1. The main results using our joint loss for weakly-supervised semantic segmentation with scribbles are shown in Sec. 5.2.

### 5.1. Normalized Cut and K-means Network

We pick a segmentation network and train with normalized cut loss only. Simpler K-means clustering loss is also experimented. We call these networks K-means Network and Normalized Cut Network.

For simplicity, we consider binary segmentation for MSRA10K saliency dataset [14], which contain simple images with good color clustering. Note that here our goal is NOT saliency segmentation, but color clustering using neural networks. We set $\sigma_{rgb} = 15$ and $\sigma_{xy} = 40$ for normalized cut and choose DeepLab-VGG-16 [13] as our network. K-means is for RGB only. We fine-tune from pre-trained saliency networks. After initialization, we train the networks with clustering loss without any supervision.

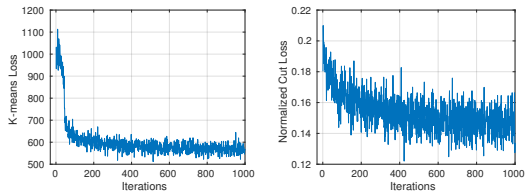

(a) Training with K-means (left) or NC (right) loss only.

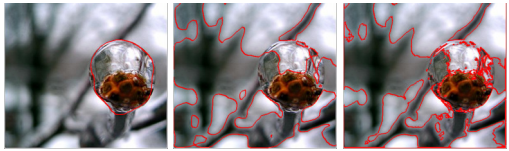

(b) K-means Network example. Left and middle: network output before and after training. Right: K-means algorithm output. Energies are 3406, 1234 and 1020 respectively.

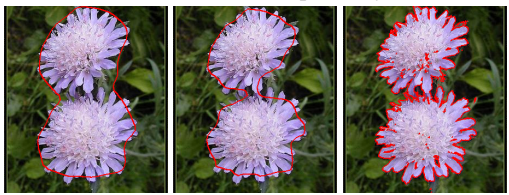

(c) Normalized Cut Network example. Left and middle: network output before and after training. Right: Using the algorithm [17] for NC. Energies are 0.18, 0.11 and 0.002 respectively.

Figure 5: Train networks to minimize K-means or Normalized Cut loss *only*. (a) loss decreases during training. In (b) and (c), color clustering improves after training, see left and middle columns. Right shows solution of standard algorithms (e.g. [17] for NC) for such clustering objectives.

Fig. 4 shows visualization of normalized cut gradients (w.r.t. softmax input) during training. It can be seen that such gradients drives segmentation towards better color clustering, particularly along object boundaries, which are more likely to be misclassified.

Indeed, we found neural networks can minimize K-means clustering or normalized cut loss by a large margin, see training loss decrease in Fig. 5a. Sample outputs of our K-means Network and Normalized Cut Network are shown in the middle column of Fig. 5b 5c. After training, the networks give segmentations of better color clustering. We also compare our *parametric network* to standard *algorithm* for such objectives, e.g. iterative K-means algorithm for K-means objective and [17] for Normalized Cut.

## 5.2. Weakly-supervised Semantic Segmentation

In this section, we first describe experiment setup, baseline and implementation details. Results with our partial cross entropy loss and joint loss are given in Sec. 5.2.1. We compare to segmentation proposals based approaches in Sec. 5.2.2. Training with our joint loss gives the best se-mantic segmentation by scribble supervision. Training with partial cross entropy loss only works surprisingly well, even better than training from GrabCut proposals. Lastly in Sec. 5.2.3, we train other networks, e.g. DeepLab-ResNet-101 [13], to show general applicability of our framework.

| loss | | | MSC | CRF | mIOU (%) |
|---|---|---|---|---|---|
| pCE | NEL | NC | | | |
| ✓ | | | | | 55.8 |
| ✓ | ✓ | | | | 56.1 |
| ✓ | ✓ | ✓ | | | 59.7 |
| ✓ | ✓ | ✓ | ✓ | | 60.5 |
| ✓ | ✓ | ✓ | ✓ | ✓ | **65.1** |

Table 1: Effect of different losses, including partial cross entropy (pCE), normalized cut (NC) and non-exist label (NEL) penalty. With our joint loss, deeplab-MSc-largeFOV-CRF gives 65.1% on PASCAL VOC 2012 *val* set.

**Dataset and Evaluations.** We test all methods on PASCAL VOC 2012 segmentation benchmark [18], which has 10,582 images in its augmented training set [24] and 1,449 images for validation. As convention, mIOU (intersection over union) on the validation set is reported. The training data are fully annotated masks for full supervision, or scribbles from [32] for weak supervision.

**Implementation Details.** We choose DeepLab-MSc-LargeFOV as our network architecture for direct comparison to [32]. For all networks, results before and after CRF are reported. We use the same for CRF parameters all method according to public DeepLab v2 code. Our reimplementation for the baseline with full supervision gives mIOU of 64.1% before CRF and 68.7% after post-processing.

We first train a network with partial cross entropy loss only. Then we fine-tune using extra normalized cut loss. We find such strategy to work better than directly minimizing the joint loss (6). When reporting our best results, we choose hyper-parameter $\lambda = 1.6$ and $\sigma_{rgb} = 15$, $\sigma_{xy} = 100$. For the deeper DeepLab-ResNet-101, we decrease Gaussian kernel bandwidth and use $\sigma_{rgb} = 12$, $\sigma_{xy} = 60$.

### 5.2.1 Results Using Our Loss

**Partial Cross Entropy (pCE) loss only.** With pCE only trained on scribbles, our approach with DeepLab-largeFOV gives mIOU of 55.8% (Tab. 1). Such trivial approach is overlooked in previous work. As discussed in Sec. 4.1, it can be seen as sampling of the full cross entropy loss based on scribbles. Indeed, training with pCE is even better than with segmentation proposals from GrabCut, which gives mIOU of 54.7%, see more comparison to segmentation proposal approach in Sec. 5.2.2.

For box supervision, Khoreva *et al.* [26] also used ignore region in training. With ignore region is shown to be better than with full segmentation proposal. This resonates with our argument that full proposals are misleading for training.
**Effect of Extra Normalized Cut Loss.** After training with pCE only, we fine-tune the network with extra normalized cut loss introduced in Sec. 4.2, see Fig. 6 and Tab. 1. For Tab. 1, MSC means with multi scaling branches and CRF means with post-processing. In Fig. 6, we tried different weight $\lambda$ for the extra normalized cut loss. Having this extra loss significantly boosts segmentation accuracy. We also used a loss penalizing labels not present by scribbles. The non-exist label (NEL) loss is defined as $<S_k, \mathbf{1}>$ for non-existing label $k$, penalizing its cardinality. In fact, this has similar effect as the cardinality constraints in [38]. As shown in Tab. 1, NEL loss slightly improves results. With the joint loss, we achieved mIOU of 65.1%, which is very close to full supervision (68.7%).

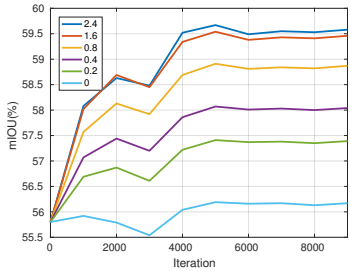

Figure 6: Effect of the weight $\lambda$ for our normalized cut loss.

| method | before CRF | after CRF |
|---|---|---|
| with Masks | 64.1 | 68.7 |
| GrabCut [42] | 55.5 | 59.7[4] |
| NormalizedCut$_{ours}$ | 58.7 | 61.3 |
| KernelCut [46] | 59.8 | 62.8 |
| ScribbleSup [32] | n/a | 63.1 |
| Ours, w/o NC loss | 56.0 | 62.0 |
| Ours, w/ NC loss | **60.5** | **65.1** |

Table 2: Our joint loss with extra normalized cut regularizer achieved the best mIOU of 65.1% with weak supervision. Training with partial cross entropy loss only is already better than through GrabCut proposals.

**Running Time for Joint Loss.** We report running time on a NVIDIA TESLA P100. For DeepLab-largeFOV with cross entropy only, it takes 0.05 sec/image for training. With extra high-order normalized cut loss, it takes 0.15 sec/image. However, we used CPU-based implementation of fast Bilateral filtering, as a subroutine for forward/backward in our normalized cut loss layer. Using existing GPU-based Bilateral filtering, e.g. for [1], will further speed up training.

### 5.2.2 Comparison to Segmentation Proposal Approach

Our framework allows direct one-pass training with scribbles, eliminating redundant iterative segmentation proposals in previous approach [32]. Here we compare to variants of segmentation proposals for training with scribbles.

There are many interactive segmentation algorithms [42, 40, 31, 46], but here we run some representative and recent ones to generate proposals, including standard GrabCut [42] and recent KernelCut [46]. Since our method includes normalized cut as a loss, we are interested in comparison to having normalized as as pre-processing (proposals). In particular, we tried a variant of [46] that only minimizes normalized cut subject to hard constraints. We denote this seeded version as NormalizedCut$_{ours}$ to differentiate from original Normalized Cut [43]. We didn't try other normalized cut approaches with scribbles, e.g. [15], since they typically rely on eigen-decomposition, which is not scalable to dense Gaussian kernel. We also compare to [32] which iterates between graph cut and training with proposals.

We train with segmentation proposals, see Fig.7 for GrabCut, KernelCut and NormalizedCut$_{ours}$. Some proposals are close to the ground truth, but most are erroneous.

As shown in Tab. 2, training with our joint loss is better than through segmentation proposals. The second competing method, ScribbleSup [32] gives mIOU that is 5.6% less than that with full supervision, which we reduce the gap to 3.6%. Note that with partial cross entropy loss only is better than training with GrabCut proposals. This confirms that networks are trained to over-fit erroneous segmentation proposals. Rather than generating unreliable proposals, we might as well train with partial cross entropy loss which is reliable. Having extra normalized loss facilitate training and significantly boost accuracy. Fig. 8 shows examples comparing joint loss and segmentation proposal approaches.

### 5.2.3 Using Different Networks

We apply our general training framework to state-of-the-art network architectures, including DeepLab-VGG16 and DeepLab-ResNet-101. For all networks, minimizing partial image    ground truth    GrabCut+FCN    NC$_{ours}$+FCN    KernelCut+FCN    with our joint loss    full supervision cross entropy loss gives descent results. Extra normalized cut regularizer consistently improves performance.

| | Full | Weak | |
|---|---|---|---|
| | | w/o NC | w/ NC |
| DeepLab-MSc-largeFOV | 64.1 | 56.0 | 60.5 |
| DeepLab-MSc-largeFOV+CRF | 68.7 | 62.0 | 65.1 |
| DeepLab-VGG16 | 68.8 | 60.4 | 62.4 |
| DeepLab-VGG16+CRF | 71.5 | 64.3 | 65.2 |
| DeepLab-ResNet101 | 75.6 | 69.5 | 72.8 |
| DeepLab-ResNet101+CRF | 76.8 | 72.8 | 74.5 |

Table 3: Training different networks with our joint loss.

---
[4] Our implementation of GrabCut is better than that reported in [32].

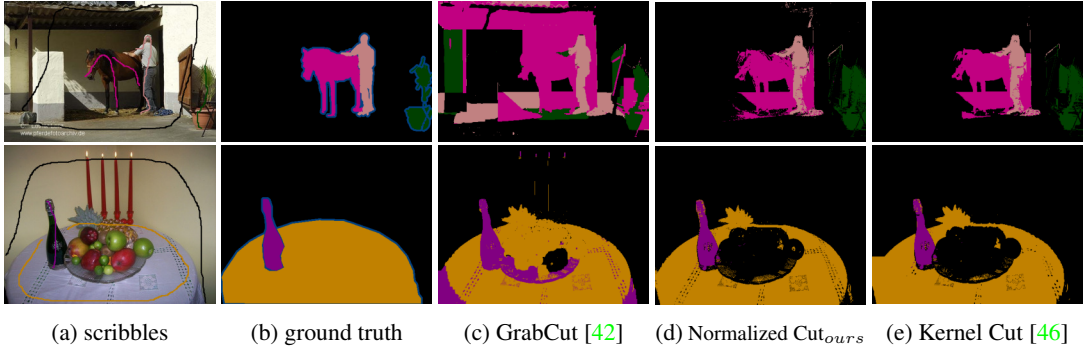

(a) scribbles    (b) ground truth    (c) GrabCut [42]    (d) Normalized Cut$_{ours}$    (e) Kernel Cut [46]

Figure 7: Proposals from interactive segmentation algorithms with seeds.

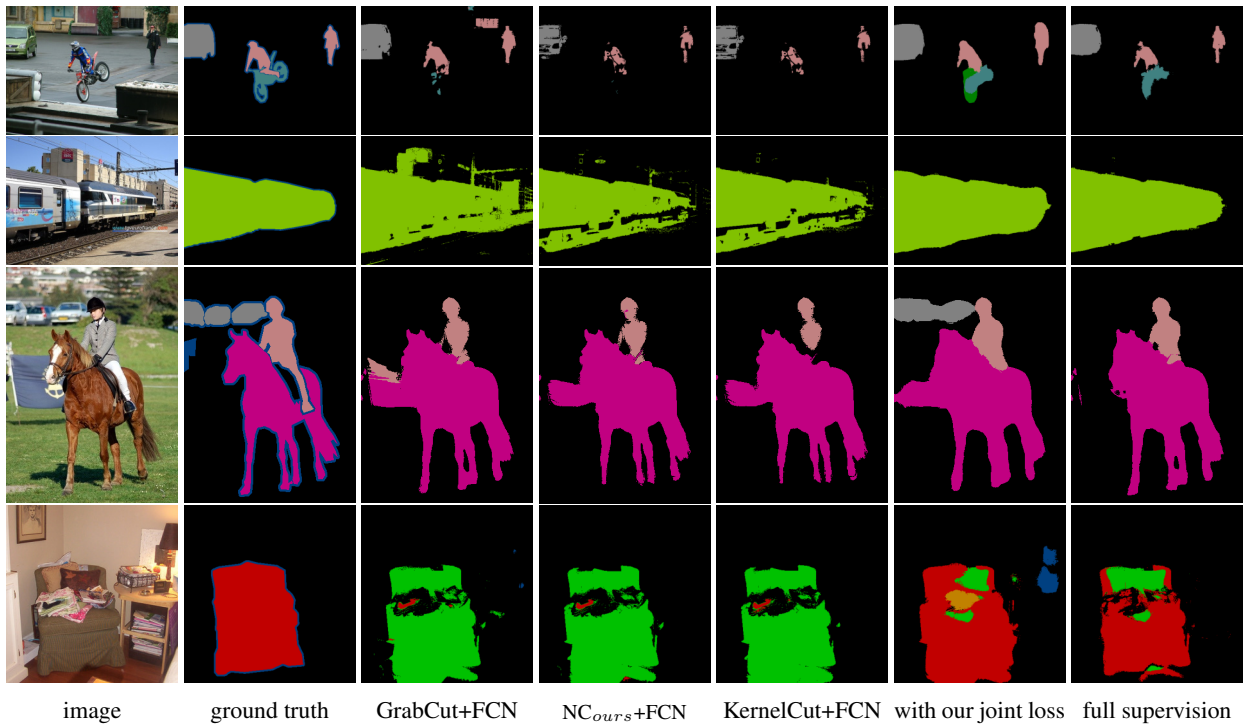

Figure 8: Testing on PASCAL VOC *val* set. Our results with scribbles are visually the closest to that with full supervision.

Training DeepLab-ResNet101 [25, 13] minimizing our joint loss gives mIOU of 74.5%, which is almost as good as that with full supervision (76.8%).

## 6. Conclusion and Future Work

We propose a novel loss for weakly-supervised segmentation with scribbles combining partial cross entropy with normalized cut. It is motivated by general ideas in semi-supervised learning and "shallow" segmentation techniques. Training with the joint loss is simpler and more principled than prior work based on iterative segmentation proposals. We show that proposal's mistakes mislead training. Even without normalized cut, our partial cross entropy loss for scribbles works surprisingly well and can be seen as loss sampling. We achieve the state-of-the-art in weakly-supervised segmentation with scribbles.

As future work, it is interesting to adjust our principled framework to other types of weak and semi supervision, e.g. with tags or boxes, and to domain adaptation. Another interesting direction is to explore MRF/CRF regularizer based losses and their different relaxations [28, 3, 46] discussed in our follow-up work [47]. Also inspired by [50], we can incorporate regularization for intermediate representation rather than for network output.

**Acknowledgments.** The authors would like to thank Ismail Ben Ayed and Olga Veksler for their feedback and comment for better presentation of this paper.